\pdfoutput=1
\documentclass{article}
\usepackage{graphicx}
\usepackage{amsfonts}
\usepackage{amsmath}

\title{Dynamic Adjacency Matrix for Video Relocalization}
\author{Yuan~Zhou, Mingfei~Wang, Ruolin~Wang, Shuwei~Huo}
\begin{document}
	\maketitle



\section{Abstract}

In this paper, we continue our work on video relocalization task. Based on using graph convolution to extract intra-video and inter-video frame features, we improve the method by using similarity-metric based graph convolution, whose weighted adjacency matrix is achieved by calculating similarity metric between features of any two different timesteps in the graph. Experiments on ActivityNet v1.2 and Thumos14 dataset shows the effectiveness of this improvement, and it outperforms the state-of-the-art methods. 

\section{Introduction}

Video retrieval has many types, due to diverse query modalities. The most traditional video retrieval methods are text query-based methods \cite{7_xu2015, 8_Mithun2018, 9_youngjae2018}, which leverage keywords or description sentences as query to search videos. Later on, image query based video retrieval methods and video query based video retrieval methods are proposed for furtherly enriching the expressive power of query modality, which use an image and a video as query respectively \cite{1_Yan2015Face, 2_Araujo2018Large, 3_Docampo2018, 10_Ye2013, 11_Song2018, 12_Chen2018, 13_zilos2017, 14_Zilos2019}. What’s more, using image and video as query modality can somewhat mitigate cross-modality issue. 

However, in practical applications, long and untrimmed videos are in common, which implies that the videos contain many complex actions, but only a few of the actions directly meet the need of users' queries. As a result, a new kind of video query task called Video Moment Retrieval (VMR) is proposed, which allows user to search for certain clips inside the video instead of the whole video. 

Like the development of video retrieval mentioned above, video moment retrieval's methods are also text-based methods at first, which aims to search the video clip that is relevant to the given text query \cite{15_Hendricks2017,16_Gao2017TALL,17_Mithun2019}. However, using text as query modality still limits the richness and complexity of the information contained in the query. Then, in order to compensate for the disadvantage of text query, video-query based video moment retrieval (video relocalization) is proposed, which is also known as Video Relocalization task \cite{18_video_reloc,19_spatial_tempoal_video_reloc}. Its aim is temporally localizing video segments in a long and untrimmed reference video, and the segment should have semantic correspondence with the given query video clip. An example of video relocalization task is shown in Figure 1. In this example, users first pick out the clip with the action of “basketball dunk” from the query video, which is the input of video relocalization task. And the task aims at retrieving a clip with the same semantic meaning in another untrimmed reference video.

For video relocalization task, the most direct way is leveraging semantic similarity between query video clip and reference video. \cite{18_video_reloc} made the first attempt to address this problem by proposing a cross-gated bilinear matching module. In their method, every timestep in the reference video is matched with all the timesteps in query video clip, thus making prediction of the starting and ending time a sequence labeling problem. \cite{ruolin} modified their feature extraction method by leveraging Attention-based Fusion Module to compute frame-level semantic similarity between query video clip and pre-extracted proposal clips in reference video. Then the generated Attention-based Fusion Tensor passes through Semantic Relevance Measurement Module to achieve the video-level semantic relevance between them, but the prediction of starting time and ending time is not related with sequence labeling. \cite{zhou2020graph} proposes Multi-Graph Feature Fusion Module, which makes the first attempt of using graph convolution in video relocalization task, and improves the evaluation metric of this task. In the article, they first treat concatenate query video feature and proposal clip feature as a graph. Then, with multiple pre-designed adjacency matrices, Multi-Graph Feature Fusion block can further fuse the feature of the two videos. 

However, the adjacency matrices are fixed, which implies that they cannot be adjusted for adapting to each query-proposal pair in training and testing stage. What’s more, the adjacency matrix should to be trained in training stage for better node connections. As a result, a video-pair dependent adjacency matrix should be built for further improve the result. 

In article \cite{my_idea_origin}, they build the adjacency matrix by measuring the similarity between features of different timesteps to improve the result in Video Temporal Action Localization task. Since each video pair has different feature matrix, and different adjacency matrices can be built. In addition, they use a fully connected layer to train for a better node feature representation so that a better adjacency matrix could be trained. We borrow their idea and put it into our work. 

Our contribution is using video-pair dependent adjacency matrix in our video relocalization task. To generate the adjacency matrix mentioned above, Weighted Graph Adjacency Matrix Generation Module is proposed. Experiments on ActivityNet v1.2 and Thumos14 dataset has proved the effectiveness of this module.

\begin{figure*}[hbt]
	\label{fig:1 vqvmr_example}
	\centering
	\includegraphics[scale=.25]{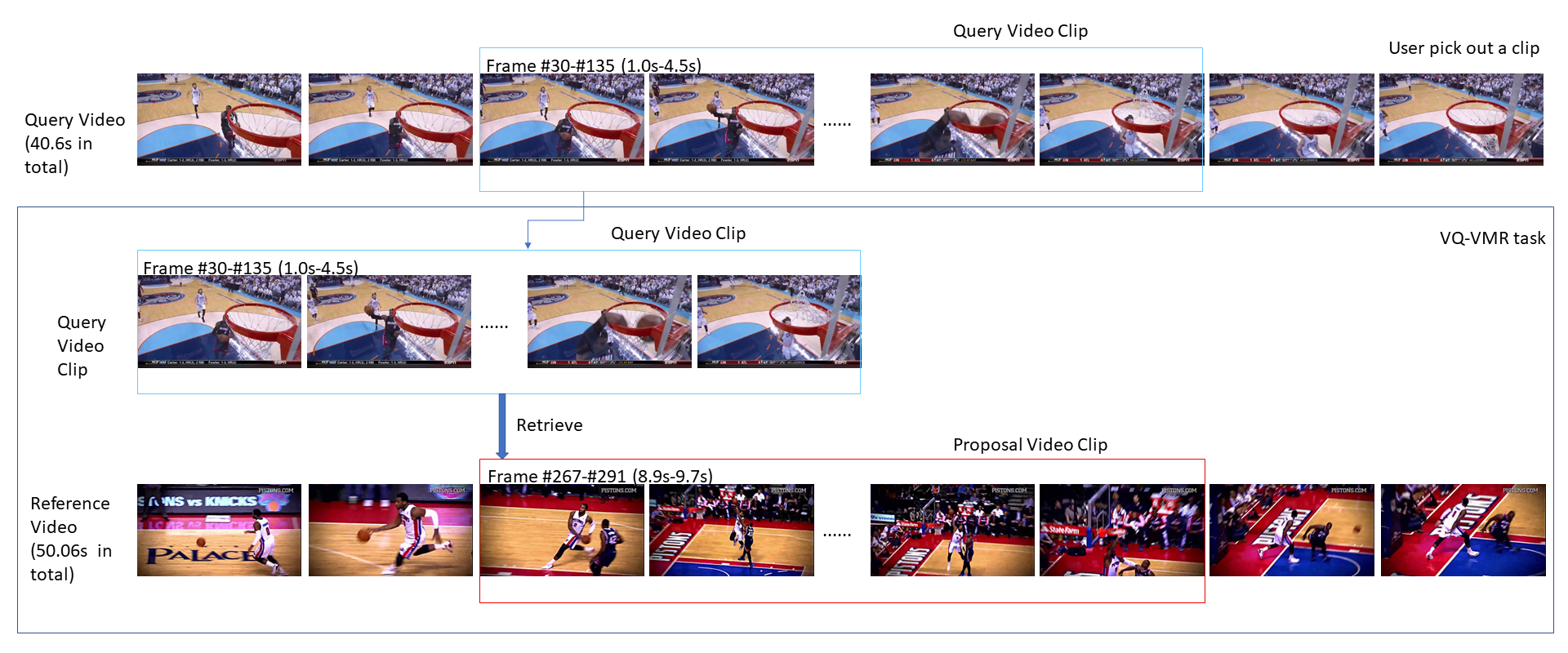}
	\caption{
		An illustration of video relocalization task: given a query video clip from a query video and an untrimmed  reference video, the task is to detect the starting point and the ending point of a segment in the reference video, which is semantically correspond to the query video. In this example, given a query video clip corresponding to “basketball dunk” (in blue box), video relocalization task aims to find out a clip which is also relevant to “basketball dunk” in another untrimmed long reference video (in red box). 
	}
\end{figure*}

\section{Related Work}

\subsection{Video Retrieval}

Video retrieval aims at selecting the video which is most relevant to the query video clip from a set of candidate videos. According to different types of query modalities, video retrieval can be divided into following categories: text-query based video retrieval, image-query based video retrieval and video-query based video retrieval. 

Text-query based video retrieval has long been tackled via joint visual-language embedding models \cite{23_Torabi2016,24_KirosSZ14,25_Hodosh2013Framing,26_Lin2014,27_Vendrov2015,28_Hu2016,29_Mao2016}. Recently, much progress has been made in this aspect. Although text and video are different modalities, which brought difficulties in studying joint feature representations, some earlier works still manage to achieve good results. Several deep video-text embedding methods \cite{23_Torabi2016,30_Xu2015Jointly,31_Otani2016Learning} has been developed by extending image-text embeddings \cite{32_Devise,33_Socher2013Grounded}. Other recent methods improve the results by utilizing concept words as semantic priors \cite{34_Yu2017}, or relying on strong representation of videos such as Recurrent Neural Network-Fisher Vector (RNN-FV) method \cite{36_KaufmanLHW16}. Also, some dominant approach leverages RNN or their variants to encode the whole multimodal sequences (e.g. \cite{23_Torabi2016,34_Yu2017,36_KaufmanLHW16}). which faces the challenge of processing cross-modality data. However, text and videos are different modalities, which means there exists inconsistency between features from the two modalities. 

Image-query based video retrieval techniques uses an image as query. Li et.al \cite{1_Yan2015Face} propose Hashing across Euclidean space and Riemannian manifold (HER) to deal with the problem of retrieving videos of a specific person given his/her face image as query. A. Araujo et.al \cite{2_Araujo2018Large} introduce a new retrieval architecture, in which the image query can be compared directly with database videos. Although compared with text modality’s feature, image modality’s feature is much more like video modality’s feature, image modality’s feature only provide appearance at one certain time, thus lacks dynamic clues.

As the expression power of text and single image are always limited, video-query based retrieval techniques are proposed to break this kind of limitation. Some video-based methods still borrow the idea of hashing from image-query based video retrieval which also map high-dimensional video features to compact binary codes so that it can address video-to-video retrieval techniques. And video retrieval has many specific applications, such as fine-grained incident video retrieval and near-duplicate video retrieval, which mainly focus on retrieving videos of the same incident and duplicated videos respectively \cite{13_zilos2017,14_Zilos2019,37_Zilos2019}.

However, in practice, videos are still very long and untrimmed. But only the clip with certain action directly meets the user’s need. To this end, video moment retrieval task is proposed, which only retrieves the video clip with certain action given a query. And our paper focuses on this task. 

\subsection{Temporal Proposal Generation}

Temporal Proposal Generation is used in Temporal Action Localization task to generate proposals of actions. Earlier proposal generation method is sliding window method, which slides the temporal window along time dimension to pick out candidates. Based on sliding windows method, \cite{38_Gaidon2013,39_Yuan2016,40_Jain2014,41_Ma2013,42_Heilbron2016} uses proposal network to predict if the current sliding window is action or not, so that some sliding windows are removed. However, sliding window method is not always satisfying, for the length of the sequence is fixed, while different actions last different time. To solve the problem, Heilbron et. al. \cite{fast_Heilbron2016Fast} propose Fast Temporal Activity Proposals method. Escorcia et. al. \cite{dap} proposes Deep Action Proposals (DAP) method to generate proposals, in which a visual encoder, a sequence encoder, a localization module and a prediction module are composited as a pipeline to extract K proposals with confidence scores over a T timestep video. Zhao et. al. \cite{49_tag_proposal} proposed a method called Temporal Actionness Grouping method. They use an actionness classifier to evaluate the binary actionness probabilities for individual snippets and a repurposed watershed algorithm to combine the snippets as proposals. In our article, we need temporal proposal generation method to generate raw proposal clip in query videos and their reference videos, and we use TAG method in \cite{49_tag_proposal} to generate our proposals. 

\subsection{Video Moment Retrieval}

Derived from video retrieval, Video Moment Retrieval needs to find out semantic-relevant clips in a video given a query. It can also be divided into two main research methods: text-query based video moment retrieval and video-query based video moment retrieval, and the latter one is also called "video relocalization" \cite{18_video_reloc}. 

Text-query based video moment retrieval focus on locating temporal segment which is the most relevant to the given text. Hendricks et. al. \cite{15_Hendricks2017} propose Moment Context Network which leverages both local and global video features over video's timesteps and effectively realize the localization in videos based on natural language queries. Gao et. al. \cite{16_Gao2017TALL} propose a Cross-Modal Temporal Regression Localizer to jointly model both textual query and video candidate moments, and its localizer outputs alignment scores between them and action's regression boundaries. With the development of attention mechanism in the field of vision and language interaction, attention is gradually used in Video Moment Retrieval models to help capturing interactions between text and video modalities. Both matching score and boundary regression are also considered in our work. We put these thoughts in our work to make it reasonable. 

Different from text-query based video moment retrieval, video-query based video moment retrieval does not have the cross-modality problem, since both query and reference are both from video modality. The methods of this part are very few. \cite{18_video_reloc} make the first attempt on by using a cross-gated bilinear matching module. In this method, the feature in reference video at every timestep is matched with every timestep in query video clip via attention mechanism. Thus, based on matching  Later on, \cite{ruolin} improved the result by using Attention-based Fusion module and Semantic Relevance Measurement module to capture frame-level relationship, however, this method still treats video relocalization task as a traditional regression problem. \cite{19_spatial_tempoal_video_reloc} extends this task to spatial-temporal level, which requires to find out both temporal segment and spatial location in the proposal video given a query video clip. In our article, our task is just video moment retrieval, which is the same as the task in \cite{18_video_reloc} and \cite{ruolin}. 

\subsection{Graph Neural Networks}

Graph Neural Networks were firstly proposed in \cite{gnn_first}, which are used for node classification, graph classification and link prediction tasks at first. 

With the success of Graph Neural Network in many aforementioned graph tasks, Graph Neural Network shows it strong power in extracting features of graph data. And many other non-graph tasks also begin using graph neural networks: they first model the input data of their tasks as graphs, and then use graph neural network to extract and fuse features. For example, \cite{gnn_image} uses graph neural networks in Image Denoising task, where a pixel is treated as a node in the graph. \cite{gnn_video} uses graph neural network in video semantic segmentation task, where each timestep is treated as a node in the graph. 

After \cite{gnn_first}, many new kinds of Graph Neural Networks are proposed, and they can be divided into two aspects: spectral based methods and spatial based methods. 

Spectral based focus on interpreting Graph Neural Networks from graph spectrums and Graph Fourier Transforms. And Laplacian matrix (which represents the graph spectrum) is calculated in this kind of methods. Graph Convolutional Network (GCN) \cite{gcn} and Graph Attention Network (GAT) \cite{gat} are two examples. 

However, spatial based methods focus on message passing from current node’s neighbors to current node, and nodes’ features are updated directly via their neighbors (and no graph spectrum information is used). GraphSAGE \cite{graphsage} is a typical example. 

For this task, we treat correlations among different timesteps as a graph data, which better represents the frame-level relationship, and we use Graph Neural Network method to fuse the feature of all timesteps in two videos. This graph modeling scheme is the same as that in \cite{gnn_video}. 

As for graph neural network methods, our method is more likely to be a spatial based method than a spectral one, for we just use the original connections among nodes in our defined graphs, and we do not utilize the spectral information of those graphs.

\section{Our Proposed Method}

In this section, we will introduce our proposed method for Video Query based Video Retrieval. The total architecture of our module can be seen in Figure 2. This section is organized as follows: Section 3.1 is problem formulation, 3.2 is an overview of our methodology, 3.3 is our proposed Weighted Graph Adjacency Matrix Generation Module for generating video-pair specific adjacency matrix. Section 3.4 is about other modules in our graph, namely graph convolution layer, score module and regression module. Section 3.5 is about losses in our method. 

\begin{figure*}[hbt]
	\label{fig:3 our proposed method}
	\centering
	\includegraphics[scale=.3]{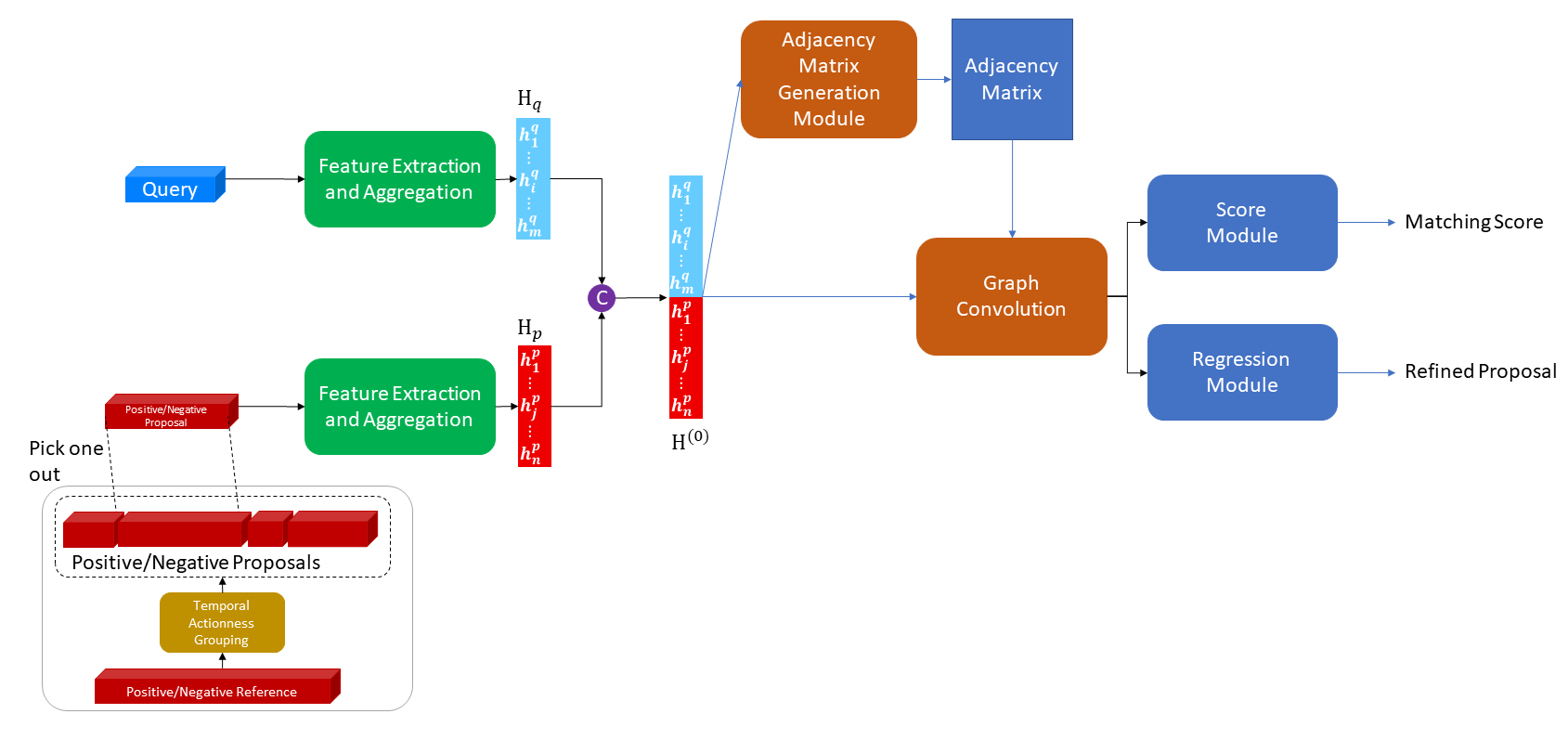}
	\caption{
	The total architecture of our module. The key is Weighted  Adjacency Matrix Generation Module, which takes the input feature matrix $H^{(0)}$ as input, and outputs an weighted adjacency matrix to represent the similarity between nodes. Different from the fixed adjacency matrix, it considers the feature similarity of the nodes inside the graph. 
	}
\end{figure*}

\subsection{Problem Formulation}

Given an query video clip Q, and a reference long video P. Our target is to get starting point and ending point $[s_{pred}, e_{pred}]$ of video clip inside P. 

To achieve this goal, in training stage, we use triplet $(q,p,n)$ as input, where $(q,p,n)$ denote query video clip, positive video (same semantic label with query) and negative video (different semantic label with query) respectively, and the total architecture of our purposed method is shown in Figure 2. This method is different from that in \cite{18_video_reloc}, where video-query based method is treated as a sequence labeling problem. We use Temporal Actionness Grouping (TAG) method to get action video clips, and for one query, we pick out one clip with same class as query as positive proposal, and one clip which has different class with query as negative proposal. Training stage aims at optimizing feature extraction module and regression module.

In testing stage, we do not use negative sample of the triplet, and only $(q,p)$ pairs are used in testing stage. For one query video clip, we also use TAG method to pick out all the proposals in the positive video. Different from picking one proposal out in training stage, we use all the proposals to predict each proposal's $[s_{pred}^{test} e_{pred}^{test}]$ as output.

\subsection{Overview of Methodology}

We follow the method in \cite{my_idea_origin}, and the only change is the generation of adjacency matrix. The overview of our method is shown as follows:

First, for each video in the input query-positive (or negative) proposal video pair, we use LSTM module to extract temporal features. Then, we concatenate the output feature of two videos at different timesteps to get a feature matrix, which is regarded as nodes’ features of a graph. To get the adjacency matrix of the graph, we pass the feature matrix into a fully connected layer to get latent feature representation, and a feature similarity metric is used to get node-wise feature similarity, and that is the way of building adjacency matrix. Then, with generated adjacency matrix and feature matrix of the graph as input, a graph convolution layer is proposed to further extract and aggregate feature. Finally, the features are sent to Score Module and Regression Module to calculate triplet loss and regression loss. 

Since the training procedure and testing procedure is almost the same with \cite{zhou2020graph}, our focus is describing the procedure of building video-dependent adjacency graph based on metric.

\subsection{Weighted Graph Adjacency Matrix Generation Module}

In \cite{zhou2020graph}, we use pre-defined weighted adjacency matrices to run the graph convolution. However, in this paper, we use similarity metric between features of two timesteps to reflect the relationship between them, which is different from the pre-defined adjacency matrix to reflect connections in different timesteps in \cite{zhou2020graph}. As a result, a video-pair specific weighted graph adjacency matrix could be built. We believe that the relationship between different timesteps should be defined by themselves. And the similarity metric between the nodes are much more reasonable than the manually-designed weights in adjacency matrix. In this work, we call it “Weighted Graph Adjacency Matrix Generation Module”. Below are the detailed building methods. And Figure 3 and Figure 4 also illustrate it. 

Following the methods in \cite{zhou2020graph}, after getting the nodes’ features $H^{(0)} \in \mathbb{R}^{2T \times d}$ via LSTM Module and feature concatenation, where $T$ denotes the number of timesteps in a video clip and $d$ denotes feature dimensions, our goal is to get an input-dependent graph adjacency matrix $\hat{A} \in \mathbb{R}^{2T\times2T}$, where each element $\hat{A}[i,j]$ shows the relationship between node $i$ and node $j$. 

First, we use a fully connected layer to learn a simple linear function $\phi$ on input feature $h_{i}^{(0)}\in\mathbb{R}^d$:
$$ \phi(h_{i}^{(0)}) = W_\phi h_{i}^{(0)} + b_\phi$$
Where $ W_\phi \in \mathbb{R}^{d^{(1)} \times d}$, $b_\phi \in \mathbb{R}^{d^{(1)}}$ are learnable weights and biases. 

This layer aims at weighting graph edges such that nodes with more similar $\phi$ have higher edge weights between them. 

Then, $\hat{A}[i,j]$ (edge weight between $\phi(h_{i}^{(0)})$ and $\phi(h_{j}^{(0)})$) can be computed as:

$$\hat{A}[i, j] = f(\phi(h_{i}^{(0)}), \phi(h_{j}^{(0)}))$$

And we use the similarity metric defined as the formula below:

$$f(h_i, h_j) = \frac{h_i^\top h_j}{\Vert h_i \Vert_2 \Vert h_i \Vert_2} $$

Also, to ensure the sparsity of our adjacency matrix, we add an L1-sparisty loss as a constraint, which will be introduced in the next subsection.  

\begin{figure*}[hbt]
	\label{fig:3 our proposed method}
	\centering
	\includegraphics[scale=.3]{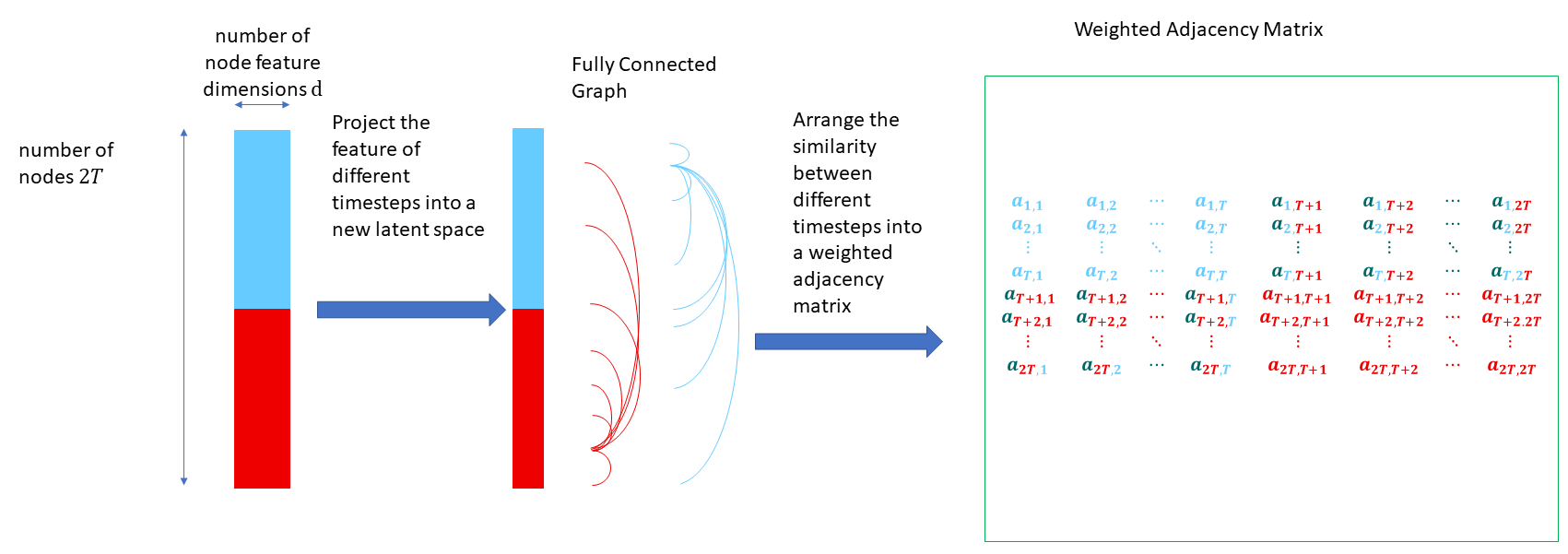}
	\caption{
	The architecture of our purposed Weighted Adjacency Matrix Generation Module. First, the concatenated feature matrix are projected into a new latent space via a Fully Connected Layer. Then, we calculated the feature similarity between different videos at different timesteps and arrange them into a weighted adjacency matrix. 
	}
\end{figure*}

\begin{figure*}[hbt]
	\label{fig:3 our proposed method}
	\centering
	\includegraphics[scale=.3]{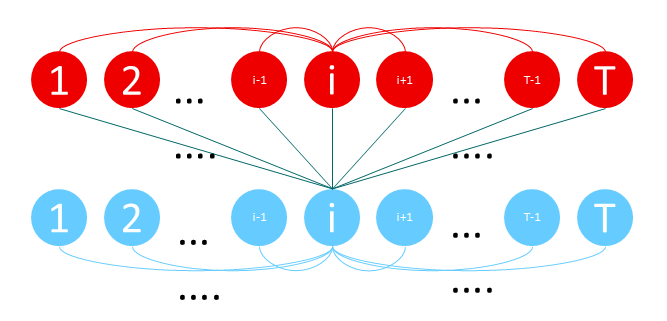}
	\caption{
	An example of our proposed graph. In this example, different from the graph in \cite{zhou2020graph}, the graph in our article is a fully-connected graph. And every edge is weighted by the similarity metric between 2 nodes. 
	}
\end{figure*}

\subsection{Other Modules in Our Method} 

In this part, we will introduce the modules after weighted adjacency matrix generation module. We will introduce graph convolution layer, score module and regression module. 

\subsubsection{Graph Convolution Layer}

After generating the weighted adjacency matrix $\hat{A}\in\mathbb{R}^{2T\times 2T}$, we take it and feature matrix $H^{(0)}\in\mathbb{R}^{2T\times d}$ as the input of graph convolution layer. 

Our Graph Convolution Layer is formulated as follows:

$$H^{(i+1)} = \sigma(\hat{A} H^{(i)} W)$$

where $\hat{A}$ is the adjacency matrix generated in the previous section, $H^{(i)}$ is the input feature matrix, $W$ is trainable weight matrix, $sigma$ is non-linear function. In the implementation of this article, different from multi-graph feature fusion method in \cite{zhou2020graph}, we only use one graph convolution layer, and the layer only contains one graph. 

\subsubsection{Score Module and Regression Module}

After passing two video’s features via a graph convolution module, a global average pooling is used for gathering the feature of the $2T$ nodes $H^{(1)}$ into one node’s feature. 

$$h_{global} = \mathrm{AvgPool}(H^{(1)}) \in \mathbb{R}^{d^{(1)}}$$

where $h_{global}$ denotes output global feature, $\mathrm{AvgPool}$ denotes average pooling, $d^{(1)}$ denotes output feature of graph convolution layer. 

\textbf{Score Module:}

$h_{global}$ is fed into this module, which consists a Multi-Layer Perception (MLP), and the module outputs an $s\in(-1,1)$ . The output $s$ is used for Triplet Loss. 

\textbf{Regression Module:}

$h_{global}$ is fed into this module, which also consists an MLP, and the module outputs regression offsets $(T_c, T_l)\in \mathbb{R}^2$, where $T_c$ and $T_l$ stand central point offset and length offset respectively. Since the proposal clip can be either too tight or too loose, this regression progress tends to find a better position.

\subsection{Loss Function}

Our loss function has 3 parts: triplet loss $L_{tri}$ is used for extracting and fusing features between query and proposals, regression loss $L_{reg}$ is used for refining starting and ending points and L1-sparsity loss $L_{L1-sparsity}$ is used for sparsening the generated weighted adjacency matrix. 

\subsubsection{Triplet Loss}

The triplet loss $L_{tri}$ is defined as follows:

$$L_{tri}=\sum_{i=1}^{N}\max(0,\gamma-S(q,p)+S(q,n))+\lambda\Vert \theta \Vert_2^2$$

where $N$ is batch size, $q,p,n$ denotes query, positive and negative video clip respectively. $\gamma$ is a hyper parameter to ensure a sufficiently large difference between the positive-query score and negative-query score. $\lambda$ is also a hyper parameter on regularization loss. In our experiment, we set $\gamma=0.5$, and $\lambda=5\times10^{-3}$, which is the same as the setting in \cite{zhou2020graph}. 

\subsubsection{Regression Loss}

The regression loss $L_{reg}$ is in the following forms:

$$L_{reg} = \frac{1}{N} \sum_{i=1}^{N} \vert T_{c,i} - T_{c,i}^* \vert + \vert T_{l,i} - T_{l,i}^* \vert$$

Where $T_{c,i}$ and $T_{l,i}$ are predicted $i^{th}$ positive proposal’s relative central and length, and $N$ is batch size. $T_{c,i}^*$ and $T_{l,i}^*$ are ground truth central points and offsets which are calculated as follows:
$$ T_{l,i}^{*} = \log( {\frac{\mathrm{len}_i}{\mathrm{len}_i^*}}) $$

$$ T_{c,i}^{*} = \frac{\mathrm{loc}_i-\mathrm{loc}_i^*}{\mathrm{len}_i^*} $$

Where $\mathrm{loc}_i$ and $\mathrm{len}_i$ denote the center coordinate and length of the $i^{th}$ proposal respectively, $\mathrm{loc}_{i}^{*}$ and $\mathrm{len}_{i}^{*}$ denote those of the corresponding ground truth segments.

\subsubsection{L1-sparsity Loss}

Based on the losses in \cite{my_idea_origin}, we add L1-sparsity loss, which is related to the generated $\hat{A}$. This loss is formulated as follows

$$L_{L1-sparsity} = \frac{1}{4T^2} \sum_{i=1}^{2T} \sum_{j=1}^{2T} \vert \hat{A}[i,j] \vert$$

This loss trains fully connected layer $\phi$ to create tighter clusters from input feature from $H^{(0)}$. 

The detailed training procedure will be shown in the experiment section. 

\subsection{Testing Stage}

After training our proposed framework, we perform the task on test set. Testing stage aims at retrieving the matching clip in an untrimmed video given a query clip. As a result, we only use the reference video which is known to have the same semantic label as query video clip, and no “negative” video is used in test stage. The query video clip and reference video mentioned above are paired to be our input in test stage.

There are two procedures in testing stage: proposal selection and proposal refinement.

\subsubsection{Proposal selection}

Given a query video clip $q$, we first get M proposals of the reference video using TAG method. Then, we calculate the score between query video clip and each of the M proposals. And the proposal with highest score is selected, which can be expressed as

$$m=\arg \max_m S(q, p_m)$$

\subsubsection{Proposal Refinement} 

After selecting the proposal with highest score, whose index is $m$, the boundary of the $m^{th}$ proposal is then refined based on the regression module in Figure 3. 

In Regression Module, we have:

$$T_{l,i}^*=\log ( {\frac{\mathrm{len}_{i}}{\mathrm{len}_{i}^*}})$$

$$T_{c,i}^*=\frac{\mathrm{loc}_{i}-\mathrm{loc}_{i}^*}{\mathrm{len}_{i}^*}$$

Where ${\mathrm loc}_{i}$ and ${\mathrm len}_{i}$ are the predicted center coordinate and length of the $i^{th}$ proposal respectively, and $\mathrm{loc}_{i}^{\ast}$ and $\mathrm{len}_{i}^\ast$ denote those of the corresponding ground truth segments. In testing stage, we need to get the predicted starting and ending point, which is equivalent to solving $\mathrm{loc}_{i}^{\ast}$ and $\mathrm{len}_{i}^\ast$ with all other items already known.

Then, with refined central point and total length known, it is easy to get the refined starting and ending point. 

\section{Experiments}

In this section, to prove the effectiveness of similarity metric between different nodes, we conduct experiments on ActivityNet v1.2 dataset and Thumos14 dataset. As the results shows below, our method outperforms all the other methods in VQVMR task. 

This section is organized as follows: first, we introduce datasets and implementation details. Then we introduce our method's results with other methods' on ActivityNet v1.2 dataset and Thumos14 dataset. Finally, we show our visualization result of the generated adjacency matrix. 

\subsection{Datasets}

As for video relocalization task, \cite{18_video_reloc} first exploit and reorganize the videos in ActivityNet to form a new dataset for research. Also, we added experiments on Thumos14 dataset, which is the same as \cite{ruolin}, to prove the effectiveness of our proposed method. 

In both datasets, original videos are annotated with starting and ending point for each  action, which is referred to "ground truth" in the following paragraph.

\subsubsection{ActivityNet}

ActivityNet v1.2 \cite{21_ActivityNet} has 9682 videos, which are divided into 100 action classes. We reorganized ActivityNet v1.2 for our study. Following the split methods in \cite{18_video_reloc}, we split 80 classes, 10 classes, 10 classes for training, validation, testing respectively. In the experiment, we use the pre-extracted 500-dimension PCA features with a temporal resolution of 16. 

\subsubsection{Thumos14}

Thumos14 dataset \cite{22_THUMOS14} has many videos, but the untrimmed long videos with temporal annotations directly meet our needs. We picked out 412 of them (200 from validation data and 212 from test data in the original Thumos14 dataset) for our training and testing, which are from 20 classes. We randomly select 14 classes for training and the rest 6 classes for testing. We need to remind that two falsely annotated videos ("270" and "1496") in the testing set were excluded in the present study. 

\subsection{Implementation Details}

For both datasets, we use pretrained C3D features as input, which is the same as that in \cite{ruolin} and \cite{zhou2020graph}. PyTorch 1.4.0 is used to implement our model. Our batch size is 32, and we train our model for 64 epochs. As mentioned above, we have 3 losses, triplet loss $L_{tri}$, regression loss $L_{reg}$, and L1 graph sparsity loss $L_{L1-sparsity}$. For 3 losses, 3 separate Adam optimizers are implemented minimize the losses. For $L_{tri}$, the optimizer optimizes parameters except Regression Module and Weighted Adjacency Matrix Generation Module, and its learning rate is 1e-4. For $L_{reg}$, it only optimizes the parameters in Regression Module, and its learning rate is 1e-1. For $L_{L1-sparsity}$, it optimizes the parameters in Weighted Adjacency Matrix Generation Module, and its learning rate is 1e-2. The 3 optimizers have the same $\beta_1=0.9$ and $\beta_2=0.999$.
 
The settings of Score Module and Regression Module are shown in table, which is the same as \cite{ruolin}. 

The parameter update scheme can be summarized as follows:

In fact, we use 3 different optimizers to optimize this loss function. First, an Adam optimizer with learning\_rate=1e-2 is used to only minimize $L_{L1-sparsity}$, which is only with respect to parameters of the fully connected layer $\phi$. Then, the second Adam optimizer with learning\_rate=1e-4 is used to optimize the parameters except $\phi$ and Regression Module, which aims at minimizing $L_{tri}$. And finally, the third Adam optimizer with learning\_rate=0.1 is used to optimize the parameters of Regression Module for minimizing $L_{reg}$. 

\subsubsection{Results on ActivityNet Dataset}

The comparison result is listed in table. From the table, it is clear to see that our proposed method has advantage with the existing VQVMR methods. Comparing with \cite{18_video_reloc}, our method’s result is much higher than theirs, this is because we use pre-extracted video moment clips, and pick out the most suitable one, rather than determining whether current timestep is in the retrieval clip or not. 

Comparing with AFT+SRM method, our method’s performance is still better than it. This conclusion is firstly mentioned in \cite{zhou2020graph}, which aims at proving the effectiveness of using graph convolution. This comparison once again proves the effectiveness of using graph convolution. What’s more, comparing with in n=1, k=1 case in \cite{zhou2020graph}. The only difference between our method and “n=1, k=1” case in \cite{zhou2020graph} is that our method’s graph is not the same among all the batches. From the result we can see that using can improve the result by adjusting the weighted adjacency matrix adaptively according to the input video pair and feature similarity metric. 

We also conduct experiments on different kinds of graphs, for example, “n=1, k=1\&2” and “n=1, k=1\&2\&3” in \cite{zhou2020graph}. The results also show the effectiveness of using metrics between timesteps to build graph adjacency matrix. 

\begin{table}
	\centering
	\caption{Results on ActivityNet Dataset}
	\begin{tabular}{|l|l|l|l|l|l|l|l|l|}
		\hline
		Methods$\backslash$tIoU & 0.5 & 0.6 & 0.7 & 0.75 & 0.8 & 0.85 & 0.9 & 0.95 \\ \hline
		Chance & 0.161 & 0.113 & 0.056 & - & 0.031 & - & 0.012 & - \\ \hline
		Frame-Level Baseline & 0.202 & 0.146 & 0.104 & - & 0.054 & - & 0.025 & - \\ \hline
		Video-Level Baseline & 0.254 & 0.181 & 0.127 & - & 0.063 & - & 0.026 & - \\ \hline
		SST & 0.347 & 0.258 & 0.183 & - & 0.081 & - & 0.03 & - \\ \hline
		Cross-Gated Bilinear Matching & 0.458 & 0.377 & 0.282 & - & 0.171 & - & 0.073 & - \\ \hline
		AFT+SRM & 0.6355 & 0.6346 & 0.6216 & 0.5639 & 0.5276 & 0.4314 & 0.2902 & 0.1233 \\ \hline
		n=1, k=1 & 0.6333 & 0.6325 & 0.6313 & 0.5637 & 0.4747 & 0.3753 & 0.265 & 0.1162 \\ \hline
		n=1, k=1\&2 & 0.6596 & 0.6494 & 0.6452 & 0.5805 & 0.4823 & 0.3832 & 0.2711 & 0.1201 \\ \hline
		n=1, k=1\&2\&3 & 0.6766 & 0.6758 & 0.6714 & 0.5933 & 0.5043 & 0.4016 & 0.2854 & 0.1241 \\ \hline
		n=2, k=1 & 0.681 & 0.6802 & 0.6787 & 0.6023 & 0.5163 & 0.4006 & 0.2835 & 0.1245 \\ \hline
		n=2, k=1\&2 & 0.6914 & 0.6907 & 0.6877 & 0.6138 & 0.5224 & 0.4016 & 0.2846 & 0.126 \\ \hline
		n=2, k=1\&2\&3 & 0.7012 & 0.7014 & 0.6981 & 0.6309 & 0.5305 & 0.4138 & 0.2886 & 0.1291 \\ \hline
		n=3, k=1 & 0.6016 & 0.6008 & 0.597 & 0.5283 & 0.4585 & 0.3608 & 0.2489 & 0.1125 \\ \hline
		n=3, k=1\&2 & 0.6242 & 0.6238 & 0.6207 & 0.5413 & 0.4613 & 0.3483 & 0.2502 & 0.1125 \\ \hline
		n=3, k=1\&2\&3 & 0.6311 & 0.6302 & 0.6268 & 0.5448 & 0.4677 & 0.3703 & 0.2624 & 0.1136 \\ \hline
		n=1, k=1, CNN & 0.5676 & 0.5675 & 0.5641 & 0.4936 & 0.4104 & 0.3305 & 0.2323 & 0.0998 \\ \hline
		n=2, k=1\&2\&3, CNN & 0.6275 & 0.6266 & 0.6247 & 0.5539 & 0.4676 & 0.3646 & 0.2642 & 0.1192 \\ \hline
		Our Method & 0.651 & 0.6501 & 0.6465 & 0.5702 & 0.4831 & 0.3798 & 0.266 & 0.1247 \\ \hline
	\end{tabular}

\end{table}

\subsubsection{Results on Thumos14 dataset}

Like \cite{ruolin} and \cite{zhou2020graph}, to further prove the effectiveness of our proposed method, we also conduct experiments on Thumos14 dataset. The experimental settings of Thuoms14 is mentioned above, which is the same as that in \cite{ruolin} and \cite{zhou2020graph}. 

The results are listed in Table . From the table, it is easy to show that the pattern in ActivityNet dataset still appears in Thumos14 dataset (when tIoU gets larger, the mAP descends) We also find that all the “graph-convolution based” methods is better than AFT+SRM methods, which is different from that in ActivityNet. We conclude this for disparity of the two datasets. Of all the methods, our proposed method is the best, and mAP is higher than other graph-convolution based methods, which also shows the power of using our method. And our method’s result is better than the AFT+SRM methods in \cite{ruolin}. 

\begin{table}
	\centering
	\caption{Results on Thumos14 Dataset}
	\begin{tabular}{|l|l|l|l|l|l|}
		\hline
		Methods$\backslash$tIoU & 0.5 & 0.6 & 0.7 & 0.8 & 0.9 \\ \hline
		Chance &  &  &  &  &  \\ \hline
		Frame-Level Baseline &  &  &  &  &  \\ \hline
		Video-Level Baseline &  &  &  &  &  \\ \hline
		SST &  &  &  &  &  \\ \hline
		Cross-Gated Bilinear Matching &  &  &  &  &  \\ \hline
		AFT+SRM & 0.5063 & 0.5015 & 0.4797 & 0.3133 & 0.1206 \\ \hline
		n=1, k=1 & 0.5683 & 0.5645 & 0.539 & 0.3527 & 0.142 \\ \hline
		n=1, k=1\&2 & 0.5704 & 0.5662 & 0.5447 & 0.3525 & 0.1431 \\ \hline
		n=1, k=1\&2\&3 & 0.5783 & 0.5741 & 0.5472 & 0.3564 & 0.1444 \\ \hline
		n=2, k=1 & 0.5709 & 0.565 & 0.5263 & 0.359 & 0.1396 \\ \hline
		n=2, k=1\&2 & 0.5715 & 0.5662 & 0.5343 & 0.3613 & 0.142 \\ \hline
		n=2, k=1\&2\&3 & 0.5804 & 0.5728 & 0.5473 & 0.3724 & 0.145 \\ \hline
		n=3, k=1 & 0.5581 & 0.5493 & 0.5328 & 0.3454 & 0.1326 \\ \hline
		n=3, k=1\&2 & 0.5663 & 0.5543 & 0.533 & 0.3558 & 0.1376 \\ \hline
		n=3, k=1\&2\&3 & 0.5731 & 0.5688 & 0.5533 & 0.3606 & 0.1395 \\ \hline
		n=1, k=1, CNN & 0.4967 & 0.4915 & 0.4668 & 0.303 & 0.126 \\ \hline
		n=2, k=1\&2\&3, CNN & 0.4418 & 0.4366 & 0.4188 & 0.2568 & 0.1068 \\ \hline
		Our Method & 0.6056 & 0.601 & 0.5763 & 0.3713 & 0.145 \\ \hline
	\end{tabular}

\end{table}

\subsubsection{Qualitive Results}

Also, we show the qualitive results of our proposed method to demonstrate the effectiveness of our method intuitively. We picked out 2 classes from ActivityNet dataset (i.e., Hand Washing Clothes and Pole Vault) and two classes from Thumos14 dataset (i.e. Javelin Throw and Soccer Penalty). The results are shown in Figure . It can be seen that ground truth and our proposed method overlap a lot. (The overlap of “”, which is due to its length is relatively long.) Although the test classes and test pairs have not been seen before, it can effectively measure the semantic similarities between query and reference classes.

\begin{figure*}[hbt]
	\label{fig:3 our proposed method}
	\centering
	\includegraphics[scale=.3]{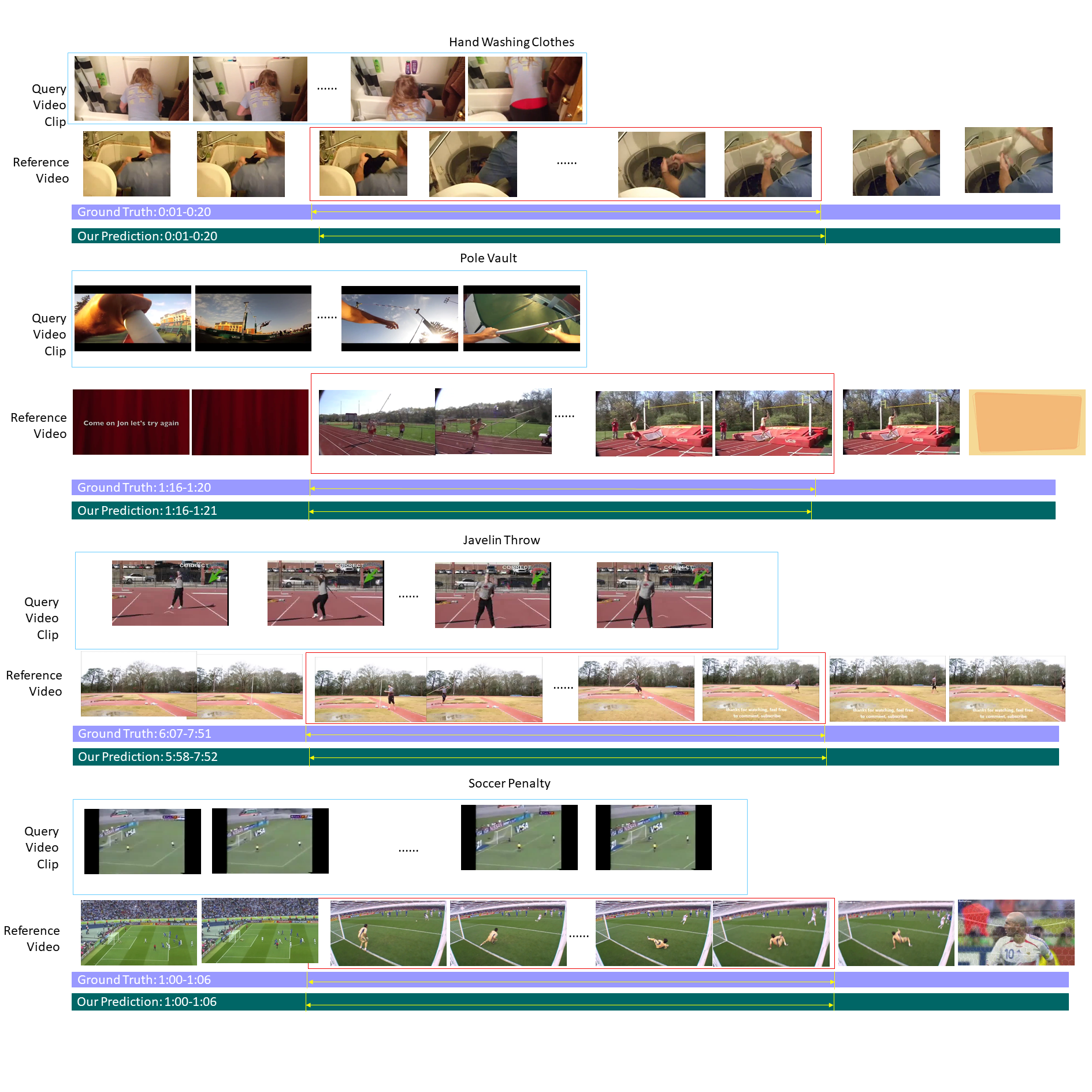}
	\caption{
Quality result of our video moment retrieval method on ActivityNet dataset and Thumos14 dataset. We pick out “Hand Washing Clothes” and “Pole Vault” from AcitivityNet, and “Javelin Throw” and “Soccer Penalty” from Thumos14 dataset. Query is the query video, refence is the video which has the same semantic label with query video. 
	}
\end{figure*}

\subsection{Adjacency Matrix Visualization}

We show our adjacency matrix in Figure . From this $\mathbb{R}^{80\times80}$ matrix, we can see that the intra-video node connections are strengthened. The inter-video node connection’s weights are not as big as the intra-video node connections, but some values are relatively higher than the others, which implies the nodes from 2 video clips are more similar, and the connection is somewhat important among all the inter-video node connections.

We also posted the pre-designed adjacency matrix in \cite{zhou2020graph} in Figure . Comparing with adjacency matrix in our method, we can find that the “edge connection strategy” in these two methods are different. The connection in \cite{zhou2020graph} is focus on “inter-video part”, while our method focuses on “intra-video part”. 

\begin{figure*}[hbt]
	\label{fig:3 our proposed method}
	\centering
	\includegraphics[scale=.3]{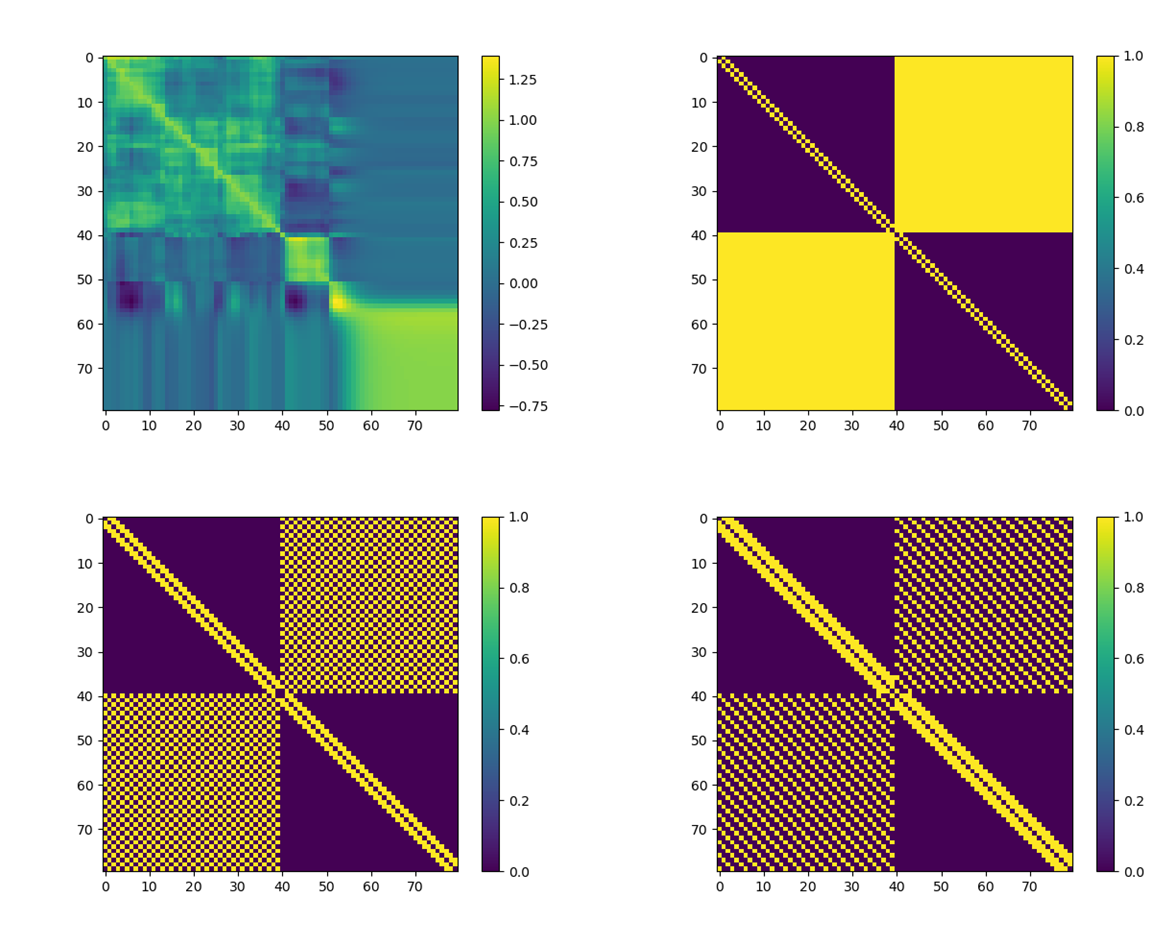}
	\caption{
	The comparison of our ”generated” adjacency matrix in this paper and pre-defined matrices in \cite{zhou2020graph}. From top left to bottom left are: the generated adjacency matrix method; k=1, k=2, k=3 adjacency matrix in \cite{zhou2020graph}.  
	}

\end{figure*}

\section{Conclusion}

In this article, we further improve video relocalization. Based on the graph convolution methods purposed before, we first concatenate the features of query and proposal to build a graph. Then we use metrics between the nodes in the graph, which is different from pre-defined adjacency weights used before. And a fully connected layer is also used for trained for Experiments on ActivityNet v1.2 and Thumos14 dataset has shown the effectiveness of our new method.

\bibliographystyle{ieeetr} 
\bibliography{my_reference} 

\end{document}